\documentclass[10pt,twocolumn,letterpaper]{article}

\usepackage{cvpr}
\usepackage{times}
\usepackage{epsfig}
\usepackage{graphicx}
\usepackage{amsmath}
\usepackage{amssymb}
\usepackage{subfigure}
\usepackage{color}
\usepackage{multirow}
\usepackage{cite}
\usepackage{epstopdf}
\usepackage{pifont}
\usepackage[dvipsnames]{xcolor}
\usepackage{caption}
\graphicspath{{figures/}}

\def\figvspace{{\vspace{-4mm}}}
\newcommand{\Paragraph}[1]{\vspace{-0mm} \noindent \textbf{#1} \hspace{0mm}}

\newcommand*{\boxedcolor}{red}
\makeatletter
\renewcommand{\boxed}[1]{\textcolor{\boxedcolor}{%
  \fbox{\normalcolor\m@th$\displaystyle#1$}}}
\makeatother

\makeatletter
  \newcommand\figcaption{\def\@captype{figure}\caption}
  \newcommand\tabcaption{\def\@captype{table}\caption}
\makeatother

\usepackage[breaklinks=true,bookmarks=true]{hyperref}

\cvprfinalcopy %

\begin{document}

\title{FSRNet: End-to-End Learning Face Super-Resolution with Facial Priors}

\author{Yu Chen$^{1}$\thanks{} ~ ~ Ying Tai$^{2}$\footnotemark[1] ~ ~ Xiaoming Liu$^3$ ~ ~ Chunhua Shen$^4$ ~ ~ Jian Yang$^1$\\
$^1$Nanjing University of Science and Technology ~ ~ ~
$^2$Youtu Lab, Tencent ~ ~ ~ \\
$^3$Michigan State University ~ ~ ~
$^4$University of Adelaide
}

\twocolumn[{
\renewcommand\twocolumn[1][]{#1}%
\maketitle
\begin{center}
    \centering
    \vspace{-6mm}
    \includegraphics[trim={0 0 0 0mm},clip,width=\textwidth]{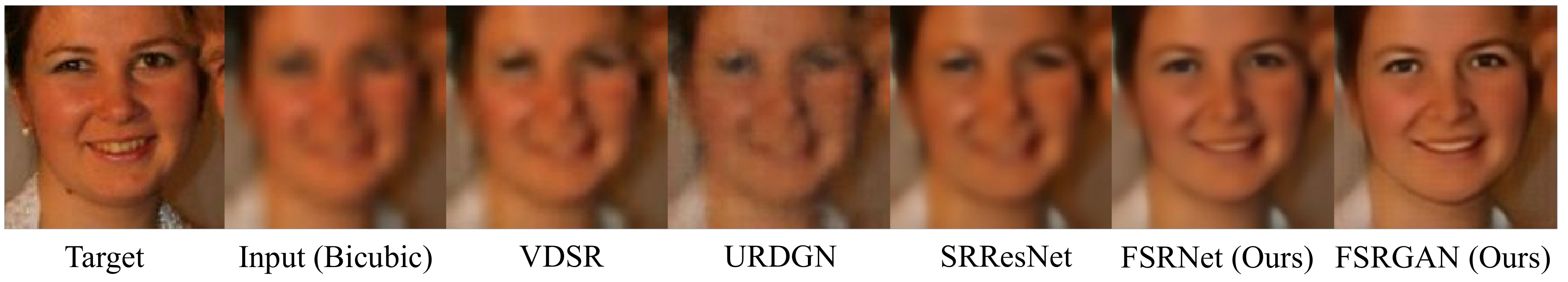} \vspace{-6mm}
    \captionof{figure}{\small Visual results of different super-resolution methods on scale factor $8$.}
    \label{fig:comp_fig1} %
\end{center}%
}]
{
\renewcommand{\thefootnote}{\fnsymbol{footnote}}
\footnotetext[1]{indicates equal contributions. This work was partially done when Yu Chen was visiting University of Adelaide.}
}

\begin{abstract}
   Face Super-Resolution (SR) is a domain-specific super-resolution problem.
   The specific facial prior knowledge could be leveraged for better super-resolving face images. %
   We present a novel deep end-to-end trainable Face Super-Resolution Network (FSRNet), which makes full use of the geometry prior, i.e., facial landmark heatmaps and parsing maps, to super-resolve very low-resolution (LR) face images without well-aligned requirement.
Specifically, we first construct a coarse SR network to recover a coarse high-resolution (HR) image.
   Then, the coarse HR image is sent to two branches: a fine SR encoder and a prior information estimation network, which extracts the image features, and estimates landmark heatmaps/parsing maps respectively.
   Both image features and prior information are sent to a fine SR decoder to recover the HR image.
   To further generate realistic faces, we propose the Face Super-Resolution Generative Adversarial Network (FSRGAN) to incorporate the adversarial loss into FSRNet.
   Moreover, we introduce two related tasks, face alignment and parsing, as the new evaluation metrics for face SR, which address the inconsistency of classic metrics w.r.t.~visual perception.
   Extensive benchmark  experiments show that FSRNet and FSRGAN significantly outperforms state of the arts for very LR face SR, both quantitatively and qualitatively.
   Code will be made available upon publication.
\end{abstract}

\section{Introduction}\label{section:1}
Face Super-Resolution (SR), a.k.a.~face hallucination, aims to generate a High-Resolution (HR) face image from a Low-Resolution (LR) input.
It is a fundamental problem in face analysis, which can greatly facilitate face-related tasks, e.g., face alignment~\cite{Amin_ICCV17,tzimiropoulos2015project}, face parsing~\cite{GFC_CVPR17}, and face recognition~\cite{NMR_PAMI16,DeepFace_CVPR14}, since most existing techniques would degrade substantially when given very LR face images.

As a special case of general image SR, there exists face-specific prior knowledge in face images, which can be pivotal for face SR and is unavailable for general image SR~\cite{MemNet_ICCV17,DRRN,SRGAN_CVPR17}.
For example, facial correspondence field could help recover accurate face shape~\cite{CBN_ECCV16}, and facial components reveal rich facial details~\cite{LCGE_IJCAI17,StructuredFH_CVPR13}.
However, as compared in Tab.~\ref{table:comparison}, the previous face SR methods that utilize facial priors all adopt multi-stage, rather than end-to-end, training strategies, which is inconvenient and complicated.

\begin{table*}[t!]
\scriptsize
\centering
  \resizebox{\linewidth}{!}{
\begin{tabular}{|c|c|c||c|c|c|c|c||c|}
\hline
\multirow{2}{*}{Method} & VDSR~\cite{VDSR_CVPR16} & SRResNet~\cite{SRGAN_CVPR17}    & StructuredFH~\cite{StructuredFH_CVPR13}       & CBN~\cite{CBN_ECCV16}   & URDGN~\cite{URDGN_ECCV16}       & AttentionFH~\cite{AttentionFH_CVPR17}      & LCGE~\cite{LCGE_IJCAI17}        & \multirow{2}{*}{FSRNet (ours)} \\
                        & (CVPR'$16$) & (CVPR'$17$) & (CVPR'$13$)   & (ECCV'$16$) & (ECCV'$16$) & (CVPR'$17$)   & (IJCAI'$17$) &  \\
\hline\hline
Facial Prior KNWL  & $\times$ & $\times$ & Components    & Dense corres.~field & $\times$ & $\times$   & Components & Landmark/parsing maps \\
\hline
Deep Model	& $\surd$ & $\surd$ & $\times$ & $\surd$ & $\surd$ & $\surd$  & $\surd$ & $\surd$ \\
\hline
End-to-End	& $\surd$ & $\surd$ & $\times$  & $\times$ & $\surd$ & $\surd$  & $\times$ & $\surd$ \\
\hline
Unaligned	& $\surd$ & $\surd$ & $\times$ & $\surd$ & $\times$  & $\times$  & $\times$ & $\surd$ \\
\hline
Scale Factor & $2/3/4$ & $2/4$ & $4$  & $2/3/4$ & $8$ & $4/8$  & $4$ & $8$ \\
\hline
\end{tabular}
}
\caption{\small Comparisons with previous state-of-the-art super-resolution methods, where VDSR and SRResNet are generic image SR methods, and StructuredFH, CBN, URDGN, AttentionFH and LCGE are face SR methods.}
\label{table:comparison}
\end{table*}

Based on deep Convolutional Neural Network (CNN), in this work, we propose a novel \textit{end-to-end trainable Face Super-Resolution Network (FSRNet)}, which estimates facial landmark heatmaps and parsing maps during training, and then uses these prior information to better super-resolve very LR face images.
It is a consensus that end-to-end training is desirable for CNN~\cite{Amin_ICCV17}, which has been validated in many areas, e.g., speech recognition~\cite{Speech_ICASSP13} and image recognition~\cite{AlexNet_NIPS12}.
Unlike previous Face SR methods that estimate local solutions in separate stages, our end-to-end framework learns the global solution directly, which is more convenient and elegant.
To be specific, since it is non-trivial to estimate facial landmarks and parsing maps directly from LR inputs, we first construct a coarse SR network to recover a coarse HR image.
Then, the coarse HR image is sent to a fine SR network, where a \textit{fine SR encoder} and a \textit{prior estimation network} share the coarse HR image as the input, followed by a \textit{fine SR decoder}.
The fine SR encoder extracts the image features, while the prior estimation network estimates landmark heatmaps and parsing maps jointly, via multi-task learning.
After that, the image features and facial prior knowledge are fed into a fine SR decoder to recover the final HR face.
The coarse and fine SR networks constitute our basic FSRNet, which already significantly outperforms the state of the arts (Fig.~\ref{fig:comp_fig1}).
To further generate realistic HR faces, \textit{Face Super-Resolution Generative Adversarial Network (FSRGAN)} is introduced to incorporate the adversarial loss into the basic FSRNet.
As in Fig.~\ref{fig:comp_fig1}, FSRGAN recovers more realistic textures than FSRNet, and clearly shows superiority over the others.

It's a consensus that Generative Adversarial Network (GAN)-based models recover visually plausible images but may suffer from low Peak Signal-to-Noise Ratio (PSNR), Structural SIMilarity (SSIM) or other quantitative metrics, while Mean Squared Error (MSE)-based deep models recover smooth images but with high PSNR/SSIM.  %
To quantitatively show the superiority of GAN-based model, in~\cite{SRGAN_CVPR17}, the authors asked $26$ users to conduct a mean opinion score testing.
However, such a testing is not objective and difficult to follow for fair comparison.
To address this problem, we introduce two related face analysis tasks, \textit{face alignment and parsing}, as the new evaluation metrics for face SR, which are demonstrated to be suitable for both MSE and GAN-based models.

In summary, the main contributions of this work include:

$\bullet$ To the best of our knowledge, this is the \textit{first} deep face super-resolution network utilizing \textit{facial geometry prior} in a convenient and elegant \textit{end-to-end training} manner.

$\bullet$ Two kinds of facial geometry priors: \textit{facial landmark heatmaps} and \textit{parsing maps} are introduced simultaneously.

$\bullet$ The proposed FSRNet achieves state-of-the-art performance when hallucinating \textit{unaligned} and \textit{very low-resolution} ($16\times 16$ pixels) face images by an upscaling factor of $8$, and the extended FSRGAN further generates more realistic face images. %

$\bullet$ Face alignment and parsing are adopted as the \textit{novel evaluation metrics} for face super-resolution, which are further demonstrated to resolve the inconsistency of classic metrics w.r.t. the visual perception.
\section{Related Work}\label{section:2}

We review the prior works from two perspectives, and contrast with the most relevant papers in Tab.~\ref{table:comparison}.

\Paragraph{Facial Prior Knowledge}
There are many face SR methods that use facial prior knowledge to better super-resolve LR faces.
Early techniques assume that faces are in a controlled setting with small variations.
Baker and Kanade~\cite{FH_Baker} proposed to learn a prior on the spatial distribution of the image gradient for frontal face images.
Wang et al.~\cite{FH_Wang} implemented the mapping between LR and HR faces by an eigen transformation.
Kolouri et al.~\cite{TBSFSR_CVPR15} learnt a nonlinear Lagrangian model for HR face images, and enhanced the degraded image by finding the model parameters that could best fit the given LR data.
Yang et al.~\cite{StructuredFH_CVPR13} incorporated the face priors by using the mapping between specific facial components.
However, the matchings between components are based on the landmark detection results that are difficult to estimate when the down-sampling factor is large.

Recently, deep convolutional neural networks have been successfully applied to the face SR task.
Zhu et al.~\cite{CBN_ECCV16} super-resolved very LR and unaligned faces in a task-alternating cascaded framework.
In their framework, face hallucination and dense correspondence field estimation are optimized alternatively.
Besides, Song et al.~\cite{LCGE_IJCAI17} proposed a two-stage method, which first generated facial components by CNNs and then synthesized fine-grained facial structures through a component enhancement method.
Different from the above methods that conduct face SR in multiple steps, our FSRNet fully leverages facial landmark heatmaps and parsing maps in an end-to-end training manner.

\begin{figure*}[t!]
  \centering
  \includegraphics[trim={0 0 0 0mm},clip,width=0.99\linewidth]{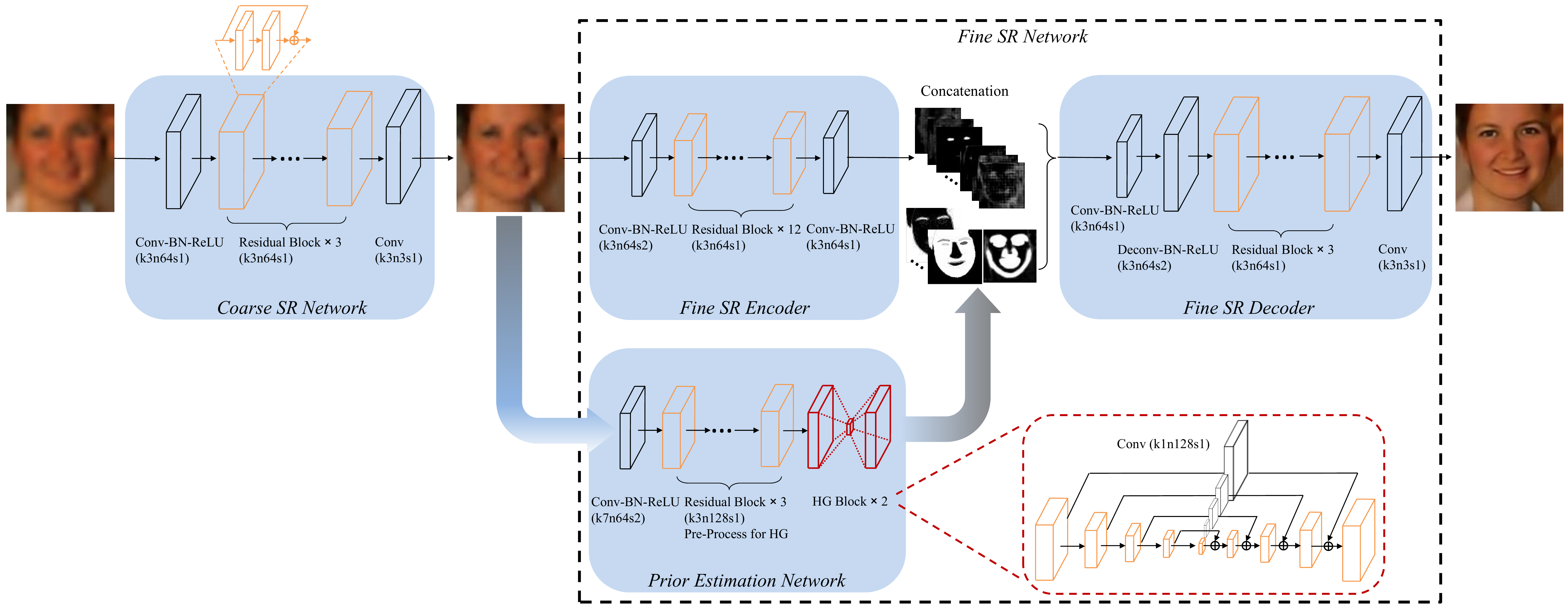}
  \caption{\small Network structure of the proposed FSRNet.
  `Conv-BN-ReLU' indicates a convolutional layer, followed by Batch Normalization (BN)~\cite{BN_ICML15} and ReLU~\cite{ReLU_ICML10}.
  `k3n64s1' indicates the kernel size to be $3\times 3$, the feature map number to be $64$ and the stride to be $1$.
  We estimate a landmark by a heatmap , but for convenience of display, we show all landmarks in one heatmap here. }
  \label{fig:network} \figvspace
\end{figure*}

\Paragraph{End-to-end Training}
End-to-end training is widely used in general image SR.
Tai et al.~\cite{DRRN} proposed Deep Recursive Residual Network (DRRN) to address the problems of model parameters and accuracy, which recursively learns the residual unit in a multi-path model.
The authors also proposed a deep end-to-end persistent memory network to address the long-term dependency problem in CNN for image restoration~\cite{MemNet_ICCV17}.
Moreover, Ledig et al.~\cite{SRGAN_CVPR17} proposed Super-Resolution Generative Adversarial Network (SRGAN) for photo-realistic image SR using a perceptual loss function that consists of an adversarial loss and a content loss.

There are also many face SR methods adopting the end-to-end training strategy.
Yu et al.~\cite{URDGN_ECCV16} investigated GAN~\cite{GAN_NIPS14} to create perceptually realistic HR face images.
The authors further proposed transformative discriminative auto-encoder to super-resolve unaligned, noisy and tiny LR face images~\cite{Yu2_CVPR17}.
More recently, Cao et al.~\cite{AttentionFH_CVPR17} proposed an attention-aware face hallucination framework, which resorts to deep reinforcement learning for sequentially discovering attended patches and then performing the facial part enhancement by fully exploiting the global image interdependency.
Different from the above methods that only rely on the power of deep models, our FSRNet is not only an end-to-end trainable Neural Network, but also combines the rich information from the facial prior knowledge.

\section{Face Super-Resolution Network}\label{section:3}
\subsection{Overview of FSRNet}
Our basic FSRNet $\mathbf{F}$ consists of four parts: \textit{coarse SR network}, \textit{fine SR encoder}, \textit{prior estimation network} and finally a \textit{fine SR decoder}.
Denote $\mathbf{x}$ as the low-resolution input image, $\mathbf{y}$ and $\mathbf{p}$ as the recovered high-resolution image and estimated prior information by FSRNet.

Since the very low-resolution input image may be too indistinct for prior estimation, we first construct the coarse SR network to recover a coarse SR image,
\begin{equation}
\label{eq:cSR}
\mathbf{y}_{c}=\mathcal{C}(\mathbf{x}) ,
\end{equation}
where $\mathcal{C}$ denotes the mapping from a LR image $\mathbf{x}$ to a coarse SR image $\mathbf{y}_{c}$ by the coarse SR network.
Then, $\mathbf{y}_{c}$ is sent to the prior estimation network $\mathcal{P}$ and fine SR encoder $\mathcal{F}$ respectively,
\begin{equation}
\label{eq:priorNet}
\mathbf{p}=\mathcal{P}(\mathbf{y}_{c}),\ \mathbf{f}=\mathcal{F}(\mathbf{y}_{c}),
\end{equation}
where $\mathbf{f}$ is the features extracted by $\mathcal{F}$.
After encoding, the SR decoder $\mathcal{D}$ is utilized to recover the SR image by \textit{concatenating} the image feature $\mathbf{f}$ and prior information $\mathbf{p}$,
\begin{equation}
\label{eq:fSR}
\mathbf{y}=\mathcal{D}(\mathbf{f},\mathbf{p}),
\end{equation}

Given a training set $\{\mathbf{x}^{(i)},\tilde{\mathbf{y}}^{(i)},\tilde{\mathbf{p}}^{(i)}\}_{i=1}^N$, where $N$ is the number of training images, $\tilde{\mathbf{y}}^{(i)}$ is the ground-truth high-resolution image of the low-resolution image $\mathbf{x}^{(i)}$ and  $\tilde{\mathbf{p}}^{(i)}$ is the corresponding ground-truth prior information, the loss function of our FSRNet is
\begin{equation}\
\small
\begin{aligned}
  \mathcal{L}_{\mathbf{F}}(\Theta) = & \frac{1}{2N}\sum_{i=1}^{N}\{\|\tilde{\mathbf{y}}^{(i)} - \mathbf{y}^{(i)}_c\|^2 + \|\tilde{\mathbf{y}}^{(i)} - \mathbf{y}^{(i)}\|^2 \\
  & + \lambda\|\tilde{\mathbf{p}}^{(i)} - \mathbf{p}^{(i)}\|^2 \},
  \end{aligned}
  \label{eq:e12}
\end{equation}
where $\Theta$ denotes the parameter set, $\lambda$ is the weight of prior loss, and $\mathbf{y}^{(i)}, \mathbf{p}^{(i)}$ are the recovered HR image and estimated prior information of the $i$-th image respectively.

\subsection{Details inside FSRNet}
We now present the details of our FSRNet, which consists of a coarse and a fine SR network, where the fine SR network contains three parts:  a prior estimation network, a fine SR encoder and a fine SR decoder.

\subsubsection{Coarse SR network}
First, we use a coarse SR network to roughly recover a coarse HR image.
The motivation is that it is non-trivial to estimate facial landmark positions and parsing maps directly from a LR input image.
Using the coarse SR network may help to ease the difficulties for estimating the priors.
The architecture of the coarse SR network is shown in Fig.~\ref{fig:network}.
It starts with a $3\times3$ convolution followed by $3$ \textit{residual blocks}~\cite{ResNet_CVPR16}.
Then another $3\times3$ convolutional layer is used to reconstruct the coarse HR image.

\subsubsection{Fine SR Network}
In the following fine SR network, the coarse HR image is sent to two branches, prior estimation network and fine encoder network, to estimate facial priors and extract features, respectively.
Then the decoder jointly uses results of both branches to recover the fine HR image.

\Paragraph{Prior Estimation Network}
Any real-world object has distinct distributions in its shape and texture, including face.
Comparing facial shape with texture, we choose to model and leverage the shape prior for two considerations.
First, when reducing the resolution from high to low, the shape information is better preserved compared to the texture, and hence is more likely to be extracted to facilitate super-resolution.
Second, it is much easier to represent shape prior than texture prior.
For example, face parsing estimates the segmentations of different face components, and  landmarks provide the accurate locations of facial keypoints.
Both represent facial shapes, while parsing carries more granularity.
In contrast, it is not clear how to represent the higher-dimensional texture prior for a specific face.

Inspired by the recent success of stacked heatmap regression in human pose estimation~\cite{Yu2017PoseNet,NewellYD16}, we adopt the HourGlass (HG) structure to estimate facial landmark heatmaps and parsing maps in our prior estimation network.
Since both priors represent the $2$D face \textit{shape}, in our prior estimation network, \textit{the features are all shared between these two tasks}, except the last layer.
The detailed structure of prior estimation network is shown in Fig.~\ref{fig:network}.
To effectively consolidate features across scales and preserve spatial information in different scales, the hourglass block uses a skip connection mechanism between symmetrical layers.
An $1\times1$ convolution layer follows to post-process the obtained features.
Finally, the shared hourglass feature is connected to two separate $1\times1$ convolution layers to generate the landmark heatmaps and the parsing maps.

\Paragraph{Fine SR Encoder}
For fine SR encoder, inspired by the success of ResNet~\cite{ResNet_CVPR16} in SR~\cite{DRRN,SRGAN_CVPR17}, we utilize the residual blocks for feature extraction.
Considering the computation cost, the size of our prior features is down-sampled to $64\times64$.
To make the feature size consistent, the fine SR encoder starts with a $3\times3$ convolutional layer of stride $2$ to down-sample the feature map to $64\times64$.
Then the ResNet structure is utilized to extract image features.

\Paragraph{Fine SR Decoder}
The fine SR decoder jointly uses the features and priors to recover the final fine HR image.
First, the prior feature $\mathbf{p}$ and image feature $\mathbf{f}$ are concatenated as the input of the decoder.
Then a $3\times3$ convolutional layer reduces the number of feature maps to $64$.
A $4\times4$ deconvolutional layer is utilized to up-sample the feature map to size $128\times128$.
Then $3$ residual blocks are used to decode the features.
Finally, a $3\times3$ convolutional layer is used to recover the fine HR image.

\subsection{FSRGAN}
As we know, GAN has shown great power in super-resolution~\cite{SRGAN_CVPR17}, which can generate photo-realistic images with superior visual effect than MSE-based deep models.
The key idea is to use a discriminative network to distinguish the super-resolved images and the real high-resolution images, and to train the SR network to deceive the discriminator.

To generate realistic high-resolution faces, our model utilizes GAN in the conditional manner~\cite{pix2pix_cvpr17}.
The objective function of the adversarial network $\mathbf{C}$ is expressed as:
\begin{equation}
\label{eq:disP}
\mathcal{L}_{\mathbf{C}}(\mathbf{F},\mathbf{C})= \mathbb{E}[\textrm{log}\mathbf{C}(\tilde{\mathbf{\mathbf{y}}},\mathbf{\mathbf{x}})]+\mathbb{E}[\textrm{log}(1-\mathbf{C}(\mathbf{F}(\mathbf{x}),\mathbf{x})],
\end{equation}
where $\mathbf{C}$ outputs the probability of the input been real and $\mathbb{E}$ is the expectation of the probability distribution.
Apart from the adversarial loss $\mathcal{L}_{\mathbf{C}}$, we further introduce a perceptual loss~\cite{PercLoss_ECCV16} using high-level feature maps (i.e., features from `relu5$\_$3' layer) of the pre-trained VGG-$16$ network~\cite{VGGNet_ICLR15} to help assess perceptually relevant characteristics,
\begin{equation}
\label{eq:feaL}
\mathcal{L}_{\mathbf{P}} = \| \phi(\mathbf{y})- \phi(\tilde{\mathbf{y}})\|^2,
\end{equation}
where $\phi$ denotes the \textit{fixed} pre-trained VGG model, and maps the images $\mathbf{y}$/$\tilde{\mathbf{y}}$ to the feature space.
In this way, the final objective function of FSRGAN is:
\begin{equation}
\label{eq:final}
\arg \min_\mathbf{F} \max_{\mathbf{C}}  \mathcal{L}_{\mathbf{F}}(\Theta)+\gamma_{\mathbf{C}}\ \mathcal{L}_{\mathbf{C}}(\mathbf{F},\mathbf{C}) 
+ \gamma_{\mathbf{P}}\ \mathcal{L}_{\mathbf{P}} , 
\end{equation}
where $\gamma_{\mathbf{C}}$ and $\gamma_{\mathbf{P}}$ are the weights of GAN and perceptual loss, respectively.

\begin{figure}[tbp]
  \centering
  \includegraphics[trim={0 0 0 0mm},clip,width=1\linewidth]{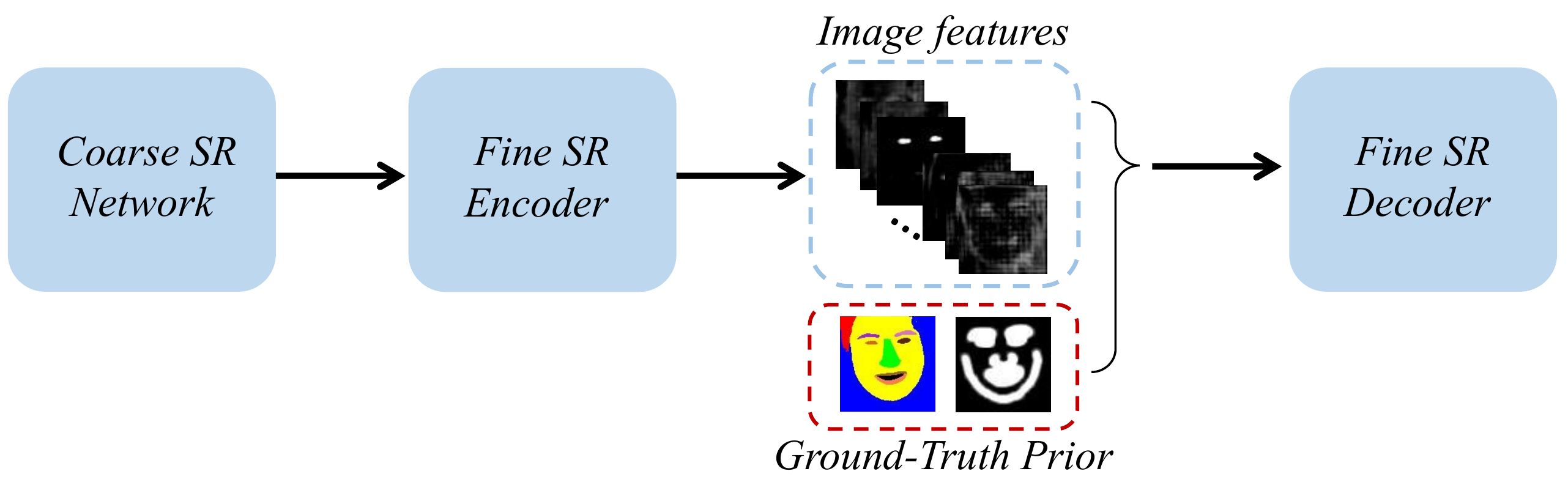}
\vspace{-5mm}
  \caption{\small Structure of ``upper-bound" model. The ground-truth priors are directly concatenated with image features.  Removing priors in the red box and increasing the number of image features by the number of channels in prior induce to the baseline model. 
  }  
  \label{fig:baseline} \figvspace
\end{figure}

\begin{figure}[tbp]
  \centering
  \includegraphics[trim={0 0 0 0mm},clip,width=0.95\linewidth]{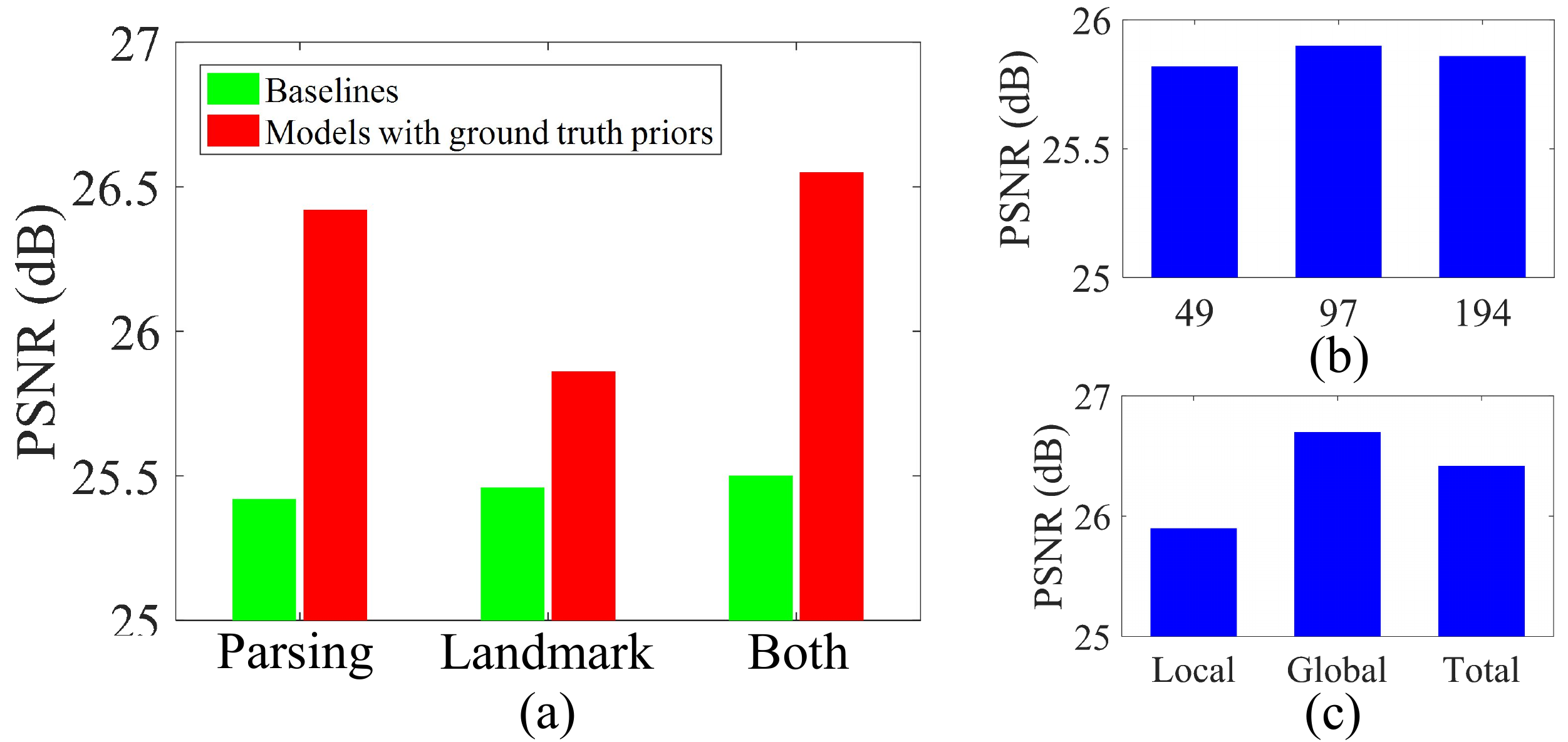}
\vspace{-2mm}
  \caption{\small Effects of facial prior knowledge.
  (a)  Comparisons between baselines and models with ground truth priors. 
The upper bound performance of landmark priors with different numbers of landmarks (b), and parsing priors with different types of parsing maps (c).}
  \label{fig:effects_of_prior} \figvspace
\end{figure}

\section{Prior Knowledge for Face Super-Resolution}\label{section:4}
In this section, we would like to answer two questions:
($1$) Is facial prior knowledge really useful for face super-resolution?
($2$) How much improvement does different facial prior knowledge bring?
To answer these questions, we conduct several tests on the $2,330$-image Helen dataset~\cite{Helen_ECCV12}.
The last $50$ images are used for testing and the others are for training.
We perform data augmentation on the training images.
Specifically, we rotate the original images by $90^\circ$, $180^\circ$, $270^\circ$ and flip them horizontally.
This results in $7$ additional augmented images for each original one.
Besides, each image in Helen dataset has a ground truth label of $194$ landmarks and $11$ parsing maps.

\Paragraph{Effects of Facial Prior Knowledge}
First, we demonstrate that facial prior knowledge is \textit{significant} for face super-resolution, even without any advanced processing steps.
We remove the prior estimation network and construct a single-branch baseline network.
Based on the baseline network, we introduce the ground truth facial prior information (i.e., landmark heatmaps and parsing maps) to the ``concatenation" layer to construct a new network, as shown in Fig.~\ref{fig:baseline}. 
For fair comparison, we keep the feature map number of ``concatenation" layer the same between two networks, which means the results can contrast the effects of the facial prior knowledge.
Fig.~\ref{fig:effects_of_prior} presents the performance of $3$ kinds of settings, including setting with or without parsing maps, landmark heatmaps, or both maps, respectively.
As we can see, the models using prior information significantly outperform the corresponding baseline models with the PSNR improvement of $0.4$ dB after using landmark heatmaps, $1.0$ dB after using parsing maps, and $1.05$ dB after using both priors, respectively.
These huge improvements on PSNR clearly signify the \textit{positive} effects of facial prior knowledge to face SR.

\begin{figure}[t!]
  \centering
  \includegraphics[trim={0 0 0 0mm},clip,width=1\linewidth]{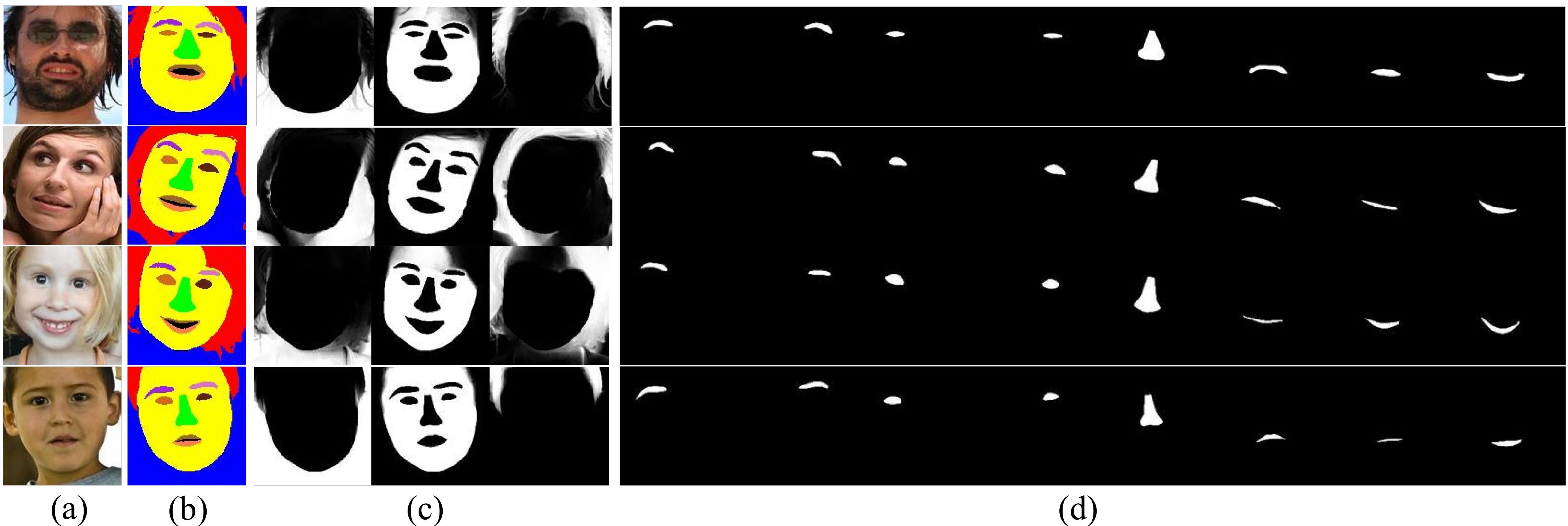}
\vspace{-5mm}
  \caption{\small Parsing maps of Helen images.
  (a) Original image.
  (b) Color visualization map generated by $11$ ground truth parsing maps~\cite{GFC_CVPR17}. It is used as part of the global parsing map.
  (c) Global parsing maps from the ground truth.
  (d) Local parsing maps from the ground truth, containing left eyebrow, right eyebrow, left eye, right eye, nose, upper lip, inner mouth, and lower lip, respectively.}
  \label{fig:parsing_maps} \figvspace
\end{figure}

\Paragraph{Upper Bound Improvements from Priors}
Next, we focus on specific prior information, and study the upper bound improvements that different priors bring.
Specifically, for facial landmarks, we introduce $3$ sets of landmarks, i.e., $49$, $97$ and $194$ landmarks, respectively.
For parsing maps, we introduce the global and local parsing maps, respectively.
The global parsing map is shown in Figs.~\ref{fig:parsing_maps}(b-c), while Fig.~\ref{fig:parsing_maps}(d) shows the local parsing maps containing different facial components.
The results of different priors are shown in Fig.~\ref{fig:effects_of_prior}.
We can observe that:
($1$) Parsing priors contain richer information for face SR and bring much more improvements than the landmark prior.
($2$) Global parsing maps are more useful than local parsing maps.
($3$) More landmark heatmaps have minor improvements than the version using $49$ landmarks.

The above results and analysis demonstrate the effects of both facial priors, and show the upper bound performance that we achieve if the priors are predicted {\it perfectly}.
Since we use the recent popular facial alignment/parsing framework as the prior estimation network, the powerful learning ability enables the network to leverage the priors as much as possible, and hence can benefit the face SR.
Apart from the benefit to PSNR, introducing facial prior may bring other advantages, such as more precise recovery of the {\it face shape}, as reflected by less errors on face alignment and parsing.
More details are presented in the next section.

\section{Experiments}\label{section:5}
\subsection{Implementation Details}\label{section:5.1}
\Paragraph{Datasets}
We conduct extensive experiments on $2$ datasets: Helen~\cite{Helen_ECCV12} and celebA~\cite{CelebA_ICCV15}. %
Experimental setting on Helen dataset is described in Sec.~\ref{section:4}.
For celebA dataset, we use the first $18,000$ images for training, and the following $100$ images for evaluation. %
It should be noted that celebA only has a ground truth of $5$ landmarks. 
We further use a recent alignment model~\cite{chen2017adversarial} to estimate the $68$ landmarks and adopt GFC~\cite{GFC_CVPR17} to estimate the parsing maps as the ground truth.

\begin{figure}[t!]
  \centering
  \includegraphics[trim={0 0 0 0mm},clip,width=0.8\linewidth]{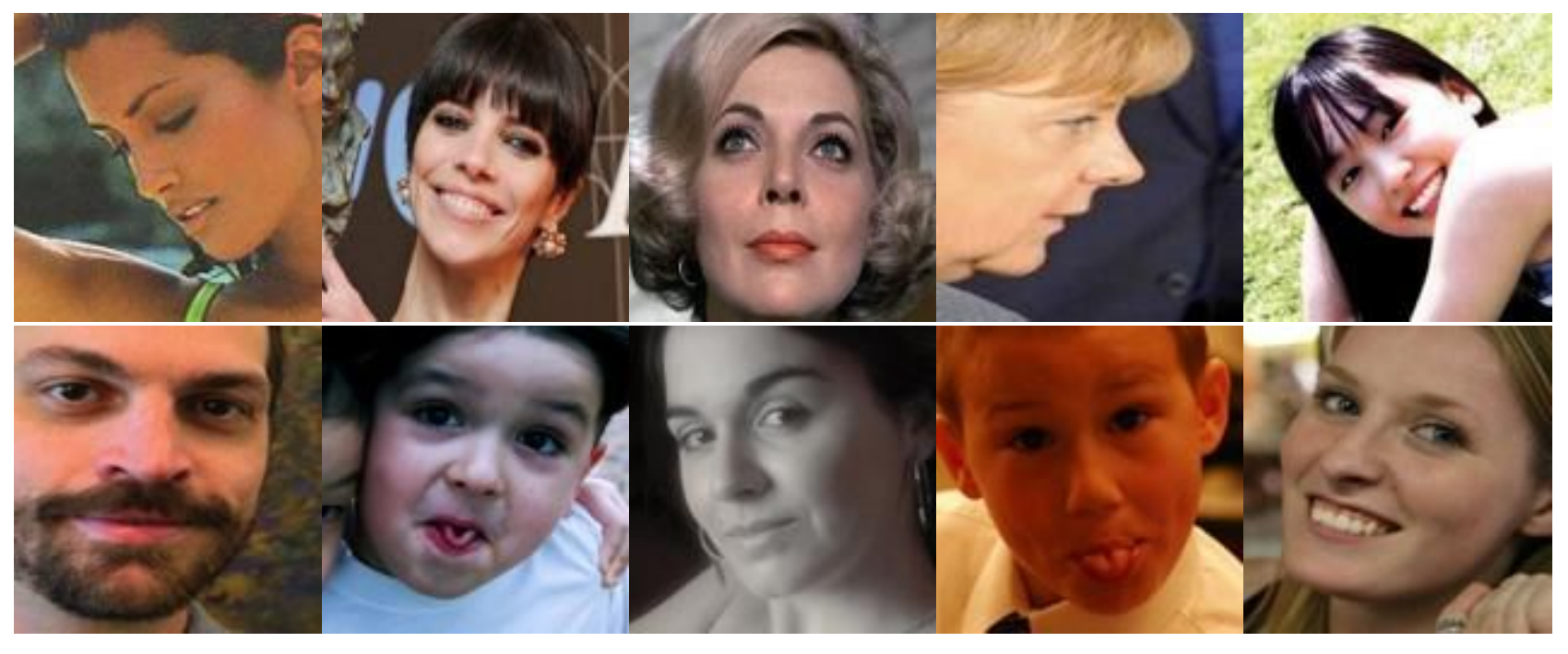}\vspace{-3mm}
  \caption{\small Training examples of CelebA (top) and Helen (bottom).}
  \label{fig:example_training} \figvspace
\end{figure}

\Paragraph{Training Setting}
We coarsely crop the training images according to their face regions and resize to $128\times 128$ without any pre-alignment operation. %
Example training images from celebA and Helen are shown in Fig.~\ref{fig:example_training}.
For testing, any popular face detector~\cite{SAFD_CVPR17} can be used to obtain the cropped image as the input.
Same as~\cite{SRGAN_CVPR17}, color images are used for training. 
The input low-resolution images are firstly enlarged by bicubic interpolation, and hence have the same size as the output high-resolution images.
For implementation, we train our model with the Torch7 toolbox~\cite{Torch7}.
The model is trained using the RMSprop algorithm with an initial learning rate of $2.5\times 10^{-4}$, and the mini-batch size of $14$.
We empirically set $\lambda=1$, $\gamma_{\mathbf{C}}=10^{-3}$ and $\gamma_{\mathbf{P}}=10^{-1}$ for both datasets.
Training a basic FSRNet on Helen dataset takes $\sim$$6$ hours on $1$ Titan X GPU.

\subsection{Ablation Study}\label{section:5.2}
\Paragraph{Effects of Estimated Priors}
We conduct ablation study on the effects of our prior estimation network.
Since our SR branch has the similar network structure as SRResNet~\cite{SRGAN_CVPR17}, we clearly show how the performance improves with different kinds of facial priors based on the performance of SRResNet.
In this test, we \textit{estimate} the facial priors through the prior estimation network instead of using the ground truth conducted in Sec.~\ref{section:4}.
Same as the tests conducted in Fig.~\ref{fig:effects_of_prior} (a), we conduct $3$ experiments to estimate the landmark heatmaps, parsing maps, or both maps, respectively.
In each experiment, we further compare our basic FSRNet with two other network structures.
Specifically, by removing the prior estimation network from our basic FSRNet, the remaining parts constitute the first network, named `Baseline$\_$v1', which has the similar structure and hence similar performance as SRResNet. %
The second network, named `Baseline$\_$v2', has the same structure as our basic FSRNet except that there is no supervision on the prior estimation network.

\begin{figure}[t!]
  \centering
  \includegraphics[trim={0 5 0 0mm},clip,width=0.8\linewidth]{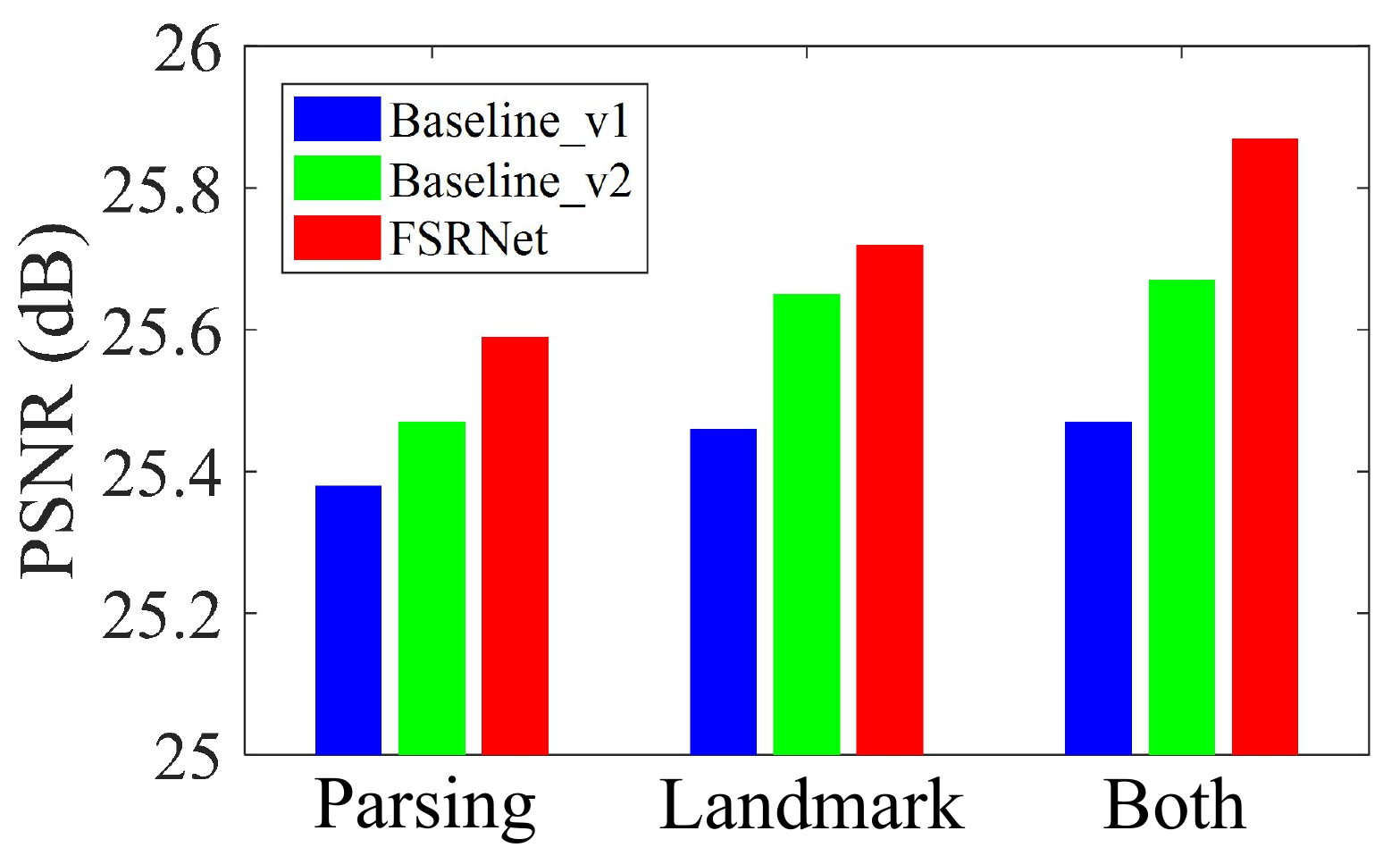}
  \vspace{-2mm}
  \caption{\small Ablation study on effects of estimated priors.
  }
  \label{fig:ablation_study} \figvspace
\end{figure}

Fig.~\ref{fig:ablation_study} shows the results of different network structures.
It can be seen that:
($1$) The second networks always outperform the first networks. The reason may be even without the supervision, the second branch learns additional features that provide more high-frequency signals to help SR.
($2$) Compared to the second networks, the supervision on prior knowledge further improves the performance, which indicates the estimated facial priors indeed have positive effects on face super-resolution.
($3$) The model using both priors achieves the best performance, which indicates richer prior information brings more improvement.
($4$) The best performance reaches $25.85$ dB, which is lower than the performance (i.e., $26.55$ dB) when using ground truth.
That means our estimated priors are not perfect and 
a better prior estimation network may result in higher model performance.

\begin{figure}[tbp]
  \begin{minipage}[b]{\linewidth}
  \centerline{\includegraphics[trim={0 -5 0 0mm},clip,width=1\linewidth]{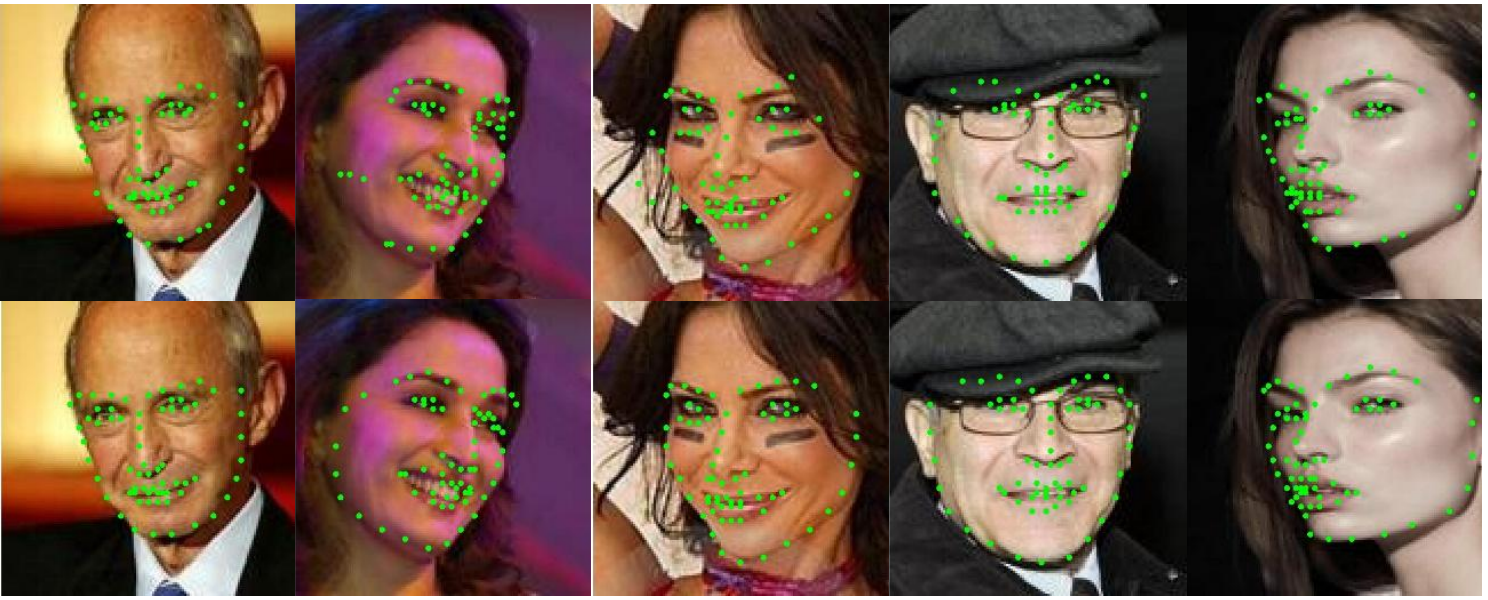}}
  \end{minipage}
  
  \begin{minipage}[b]{\linewidth}
  \centering
  \footnotesize
  \begin{tabular}{|c|c|c|c|c|c|c|}
  \hline
  CFAN &  CFSS & SDM   & DeepAlign &  FSRNet\_S1 & FSRNet\_S2 \\
  \hline
  \hline
  $9.45$  & $7.26$ & $7.88$  &\textcolor{red}{$6.50$}   & $9.44$ & \textcolor{blue}{$7.04$}  \\
  \hline
  \end{tabular}
  \end{minipage}  
  
  \caption{\small Landmark estimations by FSRNet on CelebA.
The first row shows the results of the first stacked HG (FSRNet\_S1) and the second row is of the second HG (FSRNet\_S2). Please zoom in to see the improvements.  
In the bottom, {\it NRMSEs of the first four methods are achieved by testing directly on the ground-truth HR images.}}
  \label{fig:landmark} \figvspace
\end{figure}

\Paragraph{Effects of Hourglass Numbers}
As discussed in Sec.~\ref{section:4}, a powerful prior estimation network may lead to accurate prior estimation.
Here, we study the effect of the hourglass number $h$ in the prior estimation network.
Specifically, we test $h=1/2/4$, and the PSNR results are $25.69$, $25.87$, and $25.95$ dB, respectively.
Since using more hourglasses leads to a deeper structure, the learning ability of the prior estimation network grows, and hence better performance. 
To intuitively show the adjustments in stacking more hourglasses, we show the landmark estimations of the first and second stacked hourglass in Fig~\ref{fig:landmark}. 
It can be observed that the estimation is obviously improved in the second stacking.

\begin{table*}[t!]
\scriptsize
\centering
  \resizebox{0.98\linewidth}{!}{
\begin{tabular}{|c|c|c|c|c||c|c||c|c|c|c|c|}
    \hline
    Dataset  & Bicubic & SRCNN  & VDSR & SRResNet & GLN & URDGN & FSRNet & FSRNet$\_$aug & FSRGAN \\
    \hline\hline
    Helen  & $23.69$/$0.6592$ & $23.97$/$0.6779$ & $24.61$/$0.6980$ & $25.30$/$0.7297$ &  $24.11$/$0.6922$ & $24.22$/$0.6909$  &  \textcolor{blue}{$25.87$}/\textcolor{blue}{$0.7602$} & \textcolor{red}{$26.21$}/\textcolor{red}{$0.7720$} & $25.10$/$0.7234$  \\
    \hline\hline
    celebA  & $23.75$/$0.6423$ & $24.26$/$0.6634$ & $24.83$/$0.6878$ & $25.82$/$0.7369$ &  $24.55$/$0.6867$ & $24.63$/$0.6851$  &  \textcolor{blue}{$26.31$}/\textcolor{blue}{$0.7522$} & \textcolor{red}{$26.60$}/\textcolor{red}{$0.7628$}  & $25.20$/$0.7023$  \\
    \hline
\end{tabular}
}
 \vspace{-2mm}
\caption%
{\small Benchmark super-resolution, with PSNR/SSIMs for scale factor $8$. \textcolor{red}{Red}/\textcolor{blue}{blue}  color indicate the best/second best performance.}
\label{table:benchmark_sr}\figvspace
\end{table*}

\begin{figure*}[tbp]
  \centering
  \includegraphics[trim={0 0 0 0mm},clip,width=1\linewidth]{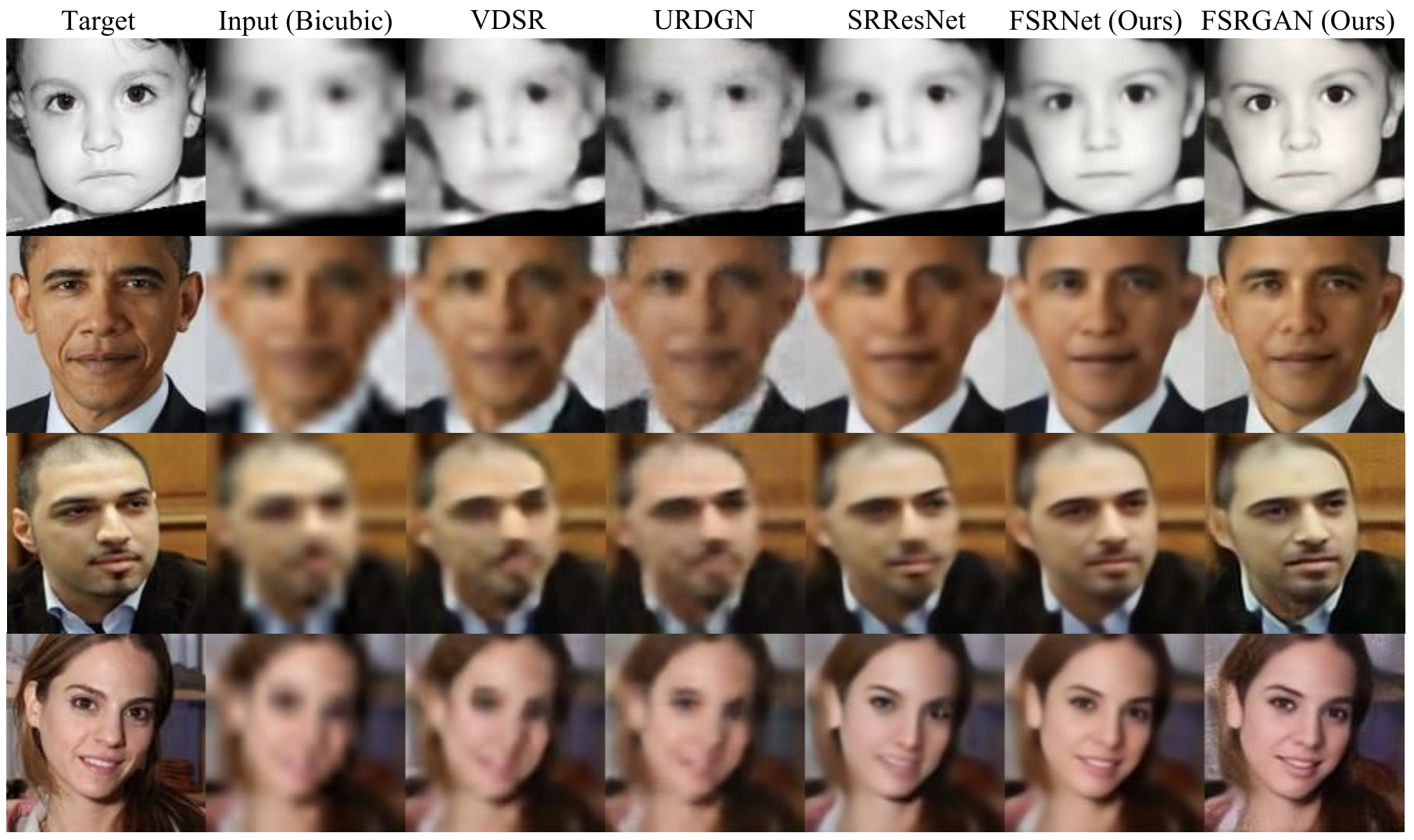}
 \vspace{-6mm}
  \caption{\small Qualitative comparisons. 
  Top two examples come from Helen and others are from celebA.
   Please zoom in to see the differences.}
  \label{fig:comp_qualitative} \figvspace
\end{figure*}

\begin{figure}[t!]
  \centering
  \includegraphics[trim={0 5 0 0mm},clip,width=1\linewidth]{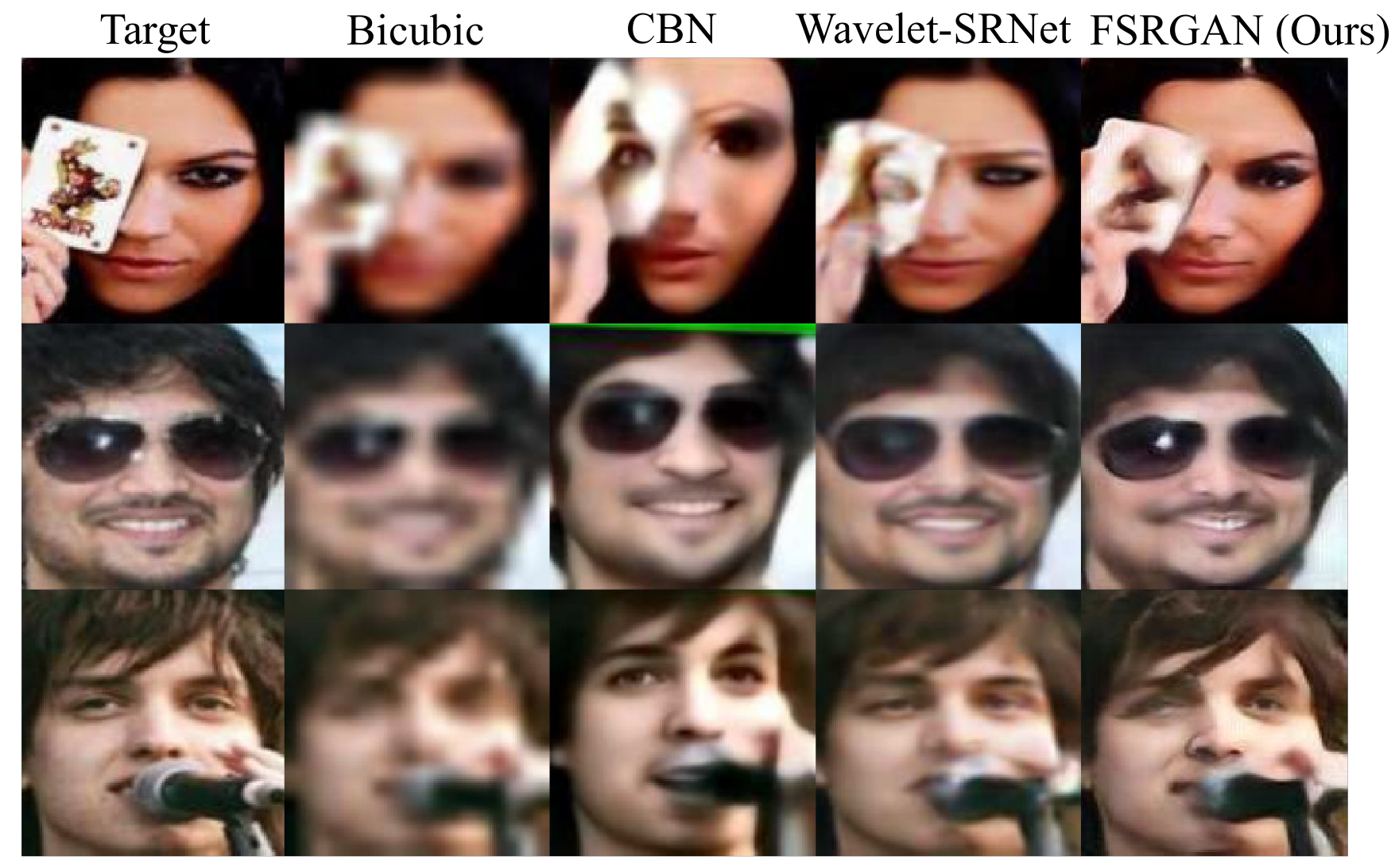}
 \vspace{-3mm}
  \caption{\small Comparisons with CBN and Wavelet-SRNet.}
  \label{fig:comp_CBN} \figvspace \vspace{-2mm}
\end{figure}

\subsection{Comparisons with State-of-the-Art Methods}\label{section:5.2}
We  compare  FSRNet with state-of-the-art SR methods, including generic SR methods like SRResNet~\cite{SRGAN_CVPR17}, VDSR~\cite{VDSR_CVPR16} and SRCNN~\cite{SRCNN_PAMI16}; and facial SR methods like GLN~\cite{GLN} and URDGN~\cite{URDGN_ECCV16}.
For fair comparison, we use the released codes of the above models and train all models with the same training set.
For URDGN~\cite{URDGN_ECCV16}, we only train the generator to report PSNR/SSIMs, but the entire GAN network for qualitative comparisons.

\Paragraph{Face Super-Resolution}
First, we compare FSRNet with the state of the arts quantitatively.
Tab.~\ref{table:benchmark_sr} summarizes quantitative results on the two datasets.
Our FSRNet significantly outperforms state of the arts in both PSNR and SSIM.
Not suprisingly, FSRGAN achieves low PSNR/SSIMs.
Besides, we also present FSRNet$\_$aug, which sends multiple augmented test images during inference and then fuse the outputs to report the results. 
This simple yet effective trick brings significant improvements.

Qualitative comparisons of FSRNet/FSRGAN with prior works are illustrated in Fig.~\ref{fig:comp_qualitative}.
Benefiting from the facial prior knowledge, our method produces relatively sharper edges and shapes, while other methods may give more blurry results.
Moreover, FSRGAN further recovers sharper facial textures than FSRNet. 

We next compare FSRGAN with two recent face SR methods: Wavelet-SRNet~\cite{waveletnet_iccv17} and CBN~\cite{CBN_ECCV16}.
We follow the same experimental setting on handling occlued face as~\cite{waveletnet_iccv17} and directly import the $16\times 16$ test examples from~\cite{waveletnet_iccv17} for super-resolving $128\times 128$ HR images.
As shown in Fig.~\ref{fig:comp_CBN}, FSRGAN achieves relatively sharper shapes (e.g., nose in all cases) than the state of the arts.

\begin{figure}[tbp]
  \centering
  \includegraphics[trim={0 0 0 0mm},clip,width=0.98\linewidth]{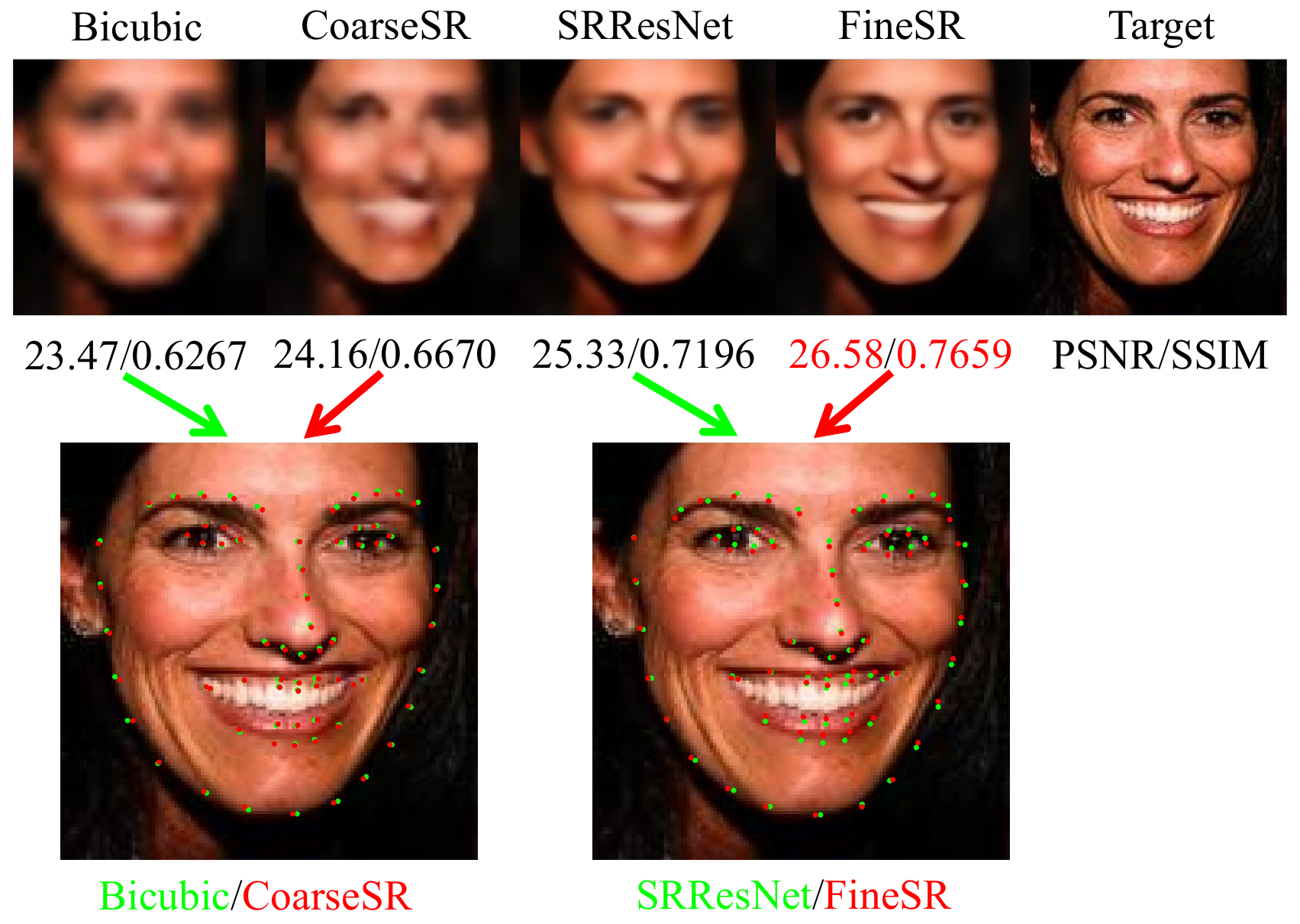}
 \vspace{-3mm}
  \caption{\small Qualitative comparison of face alignment.}
  \label{fig:alignment_eval} \figvspace
\end{figure}

\Paragraph{Face Alignment}
Apart from the evaluation of PSNR/SSIM, we introduce face alignment as a novel evaluation metric for face super-resolution, since accurate face recovery should lead to accurate shape/geometry, and hence accurate landmark points.
We adopt a popular alignment model CFAN~\cite{zhang2014coarse} to estimate the landmarks of different recovered images.
Fig.~\ref{fig:alignment_eval} shows the recovered images of SRResNet and our FSRNet, including the results from coarse SR net and final output.
Tab.~\ref{tab:comparison_face_alignmnet} presents the Normalized Root Mean Squared Error (NRMSE) results, which is a popular metric in face alignment and small NRMSE indicates better alignment performance.
From the results we can see that:
($1$) It is difficult for the state-of-the-art alignment model to estimate landmarks directly from very low-resolution images.
The estimated landmarks of the bicubic image exhibit large errors around mouth, eyes or other components.
In FSRNet, the coarse SR net can ease the alignment difficulty to some extent, which leads to better NRMSE than the input bicubic image.
($2$) Compared to SRResNet, our final output provides visually superior estimation on mouth, eyes and shape, and also achieves a large margin of $0.7$ quantitatively.
That demonstrates the effectiveness of using landmark priors for training.

On the other hand, we also compare the landmarks directly estimated by FSRNet as a by-product, with other methods~\cite{kowalski2017deep,xiong2013supervised,zhang2014coarse,zhu2015face} using their released codes, as shown in the bottom of Fig.~\ref{fig:landmark}. 
It should be noted that {\it our method starts with the LR images while others are tested directly on the ground-truth $8\times$ HR images}.
Despite the disadvantage in the input image resolution, our method outperforms most recent methods and is competitive with the state of the art.

\begin{table}[t!]
\footnotesize
\centering
\begin{tabular}{|c|c|c|c|c|c|c|c||c|c|}
\hline
  Bicubic & CoarseSR & SRResNet & FineSR & FSRGAN & Target \\
\hline
\hline
  $5.87$ & $5.42$ & $4.87$ & $4.18$  & \textcolor{red}{$3.97$} & $3.32$ \\
\hline
\end{tabular}
 \vspace{-1mm}
\caption{\small Comparisons of alignment (NRMSE) on Helen dataset.}
\label{tab:comparison_face_alignmnet} \figvspace
\end{table}

\begin{figure}[t!]
  \centering
  \includegraphics[trim={0 0 0 0mm},clip,width=1\linewidth]{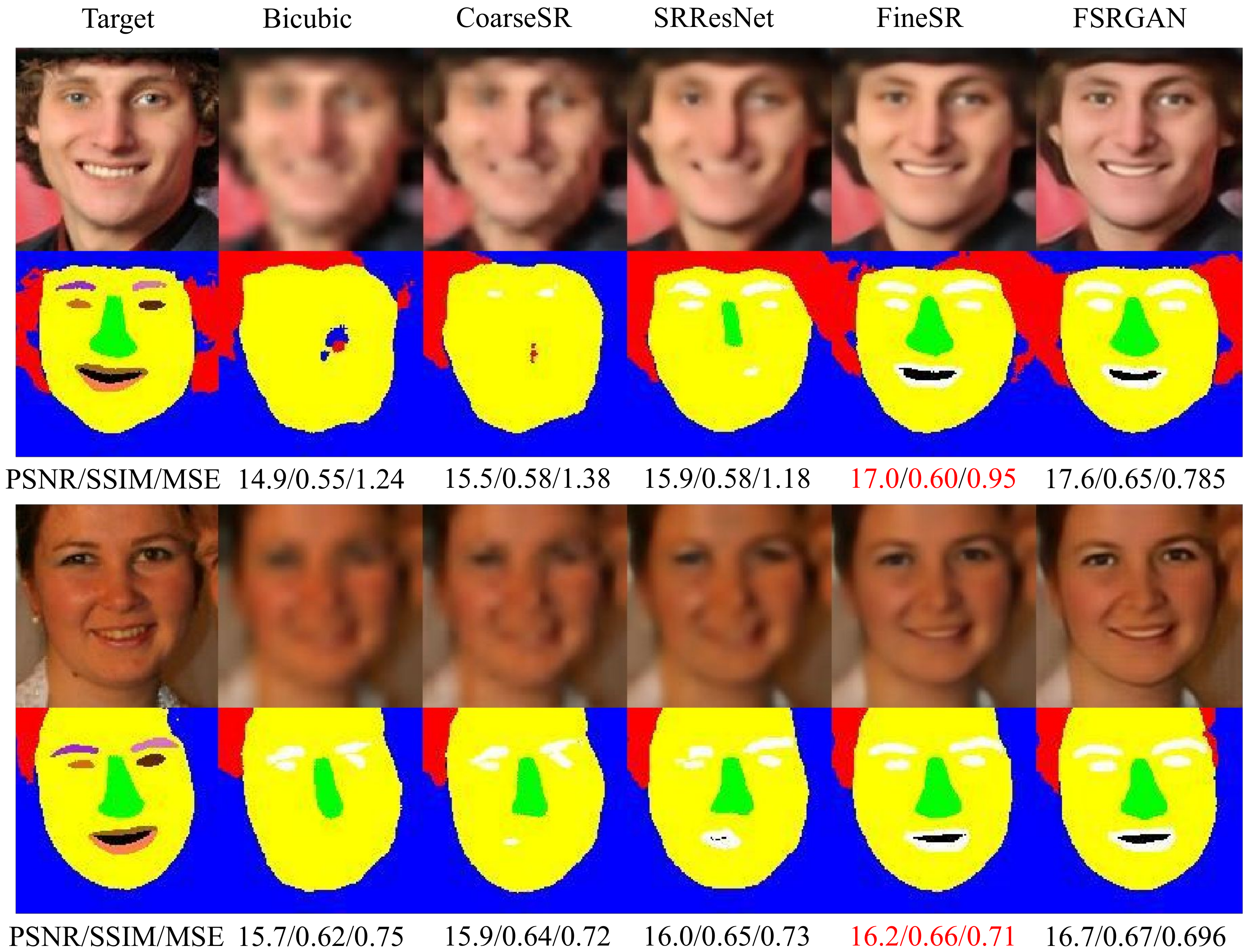}
 \vspace{-6mm}
  \caption{\small Qualitative comparison of face parsing. }
  \label{fig:parsing_eval}  %
   \vspace{-3mm}
\end{figure}

\Paragraph{Face Parsing}
We also introduce face parsing as another evaluation metric for face super-resolution.
Although our prior estimation network can predict the parsing maps from the LR inputs, for fair comparison, we adopt a recent model GFC~\cite{GFC_CVPR17} to generate the facial parsing maps for the recovered images of all methods, including the bicubic inputs, our coarse SR net, SRResNet, our fine SR net, and targets, respectively.
PSNR, SSIM, and Mean Squared Error (MSE) metrics are reported in Tab.~\ref{tab:comparison_face_parsing}.
As we can see, the coarse SR net also has positive effects on face parsing, and our FSRNet outperforms SRResNet in all of the three evaluations.
Fig.~\ref{fig:parsing_eval} presents the estimated parsing maps by~\cite{GFC_CVPR17}, the parsing maps from our final HR images recover complete and accurate components, while SRResNet may generate wrong shapes or even lose components (e.g., mouth).

Here, we adopt two side tasks, face alignment and parsing, as the new evaluation metrics for face super resolution.
They can subjectively evaluate the quality of geometry in the recovered images, which is comprementary to the classic PSNR/SSIM metrics that focus more on photometric quality.
Further, Tab.~\ref{tab:comparison_face_alignmnet} and~\ref{tab:comparison_face_parsing} show that FSRGAN outperforms FSRNet on both metrics, which is consistent with the superior visual quality in Fig.~\ref{fig:comp_qualitative}.
This consistency actually addresses one issue in GAN-based super-resolution methods, which has superior visual quality, but lower PSNR/SSIM.  
This also shows that GAN-based methods can better recover the facial geometry, in addition to perceived visual quality. 

\Paragraph{Time Complexity}
Unlike CBN that needs multiple steps and trains multiple models for face hallucination, our FSRNet is faster and more convenient to use, which only needs \textit{one forward process} for inference and costs $0.012$s on Titan X GPU, for a $128\times 128$ image.
For comparison, CBN has four cascades and totally consumes $3.84$s~\cite{CBN_ECCV16}, while the traditional face SR requires more time, e.g.,~\cite{facesr1_ijcv07} needs $8$ minutes and~\cite{Jin_CVPR15} needs $15-20$ minutes.

\begin{table}[t!]
\scriptsize
\centering
\begin{tabular}{|c|c|c|c|c|c|}
\hline
Methods  & Bicubic & CoarseSR & SRResNet  & FineSR & FSRGAN \\
\hline
\hline
PSNR  & $14.47$ & $14.91$ & $15.32$ & $15.89$ & \textcolor{red}{$16.11$} \\
\hline
SSIM  & $0.570$ & $0.585$ & $0.603$ & $0.622$ & \textcolor{red}{$0.629$}  \\
\hline
MSE  & $1.170$ & $1.100$ & $1.047$ & $0.976$ & \textcolor{red}{$0.934$}  \\
\hline
\end{tabular}
 \vspace{-1mm}
\caption{\small Comparisons of face parsing on  Helen dataset.}
\label{tab:comparison_face_parsing} \figvspace
\end{table}

\section{Conclusions}\label{section:6}
In this paper, a novel deep end-to-end trainable Face Super-Resolution Network (FSRNet) is proposed for face super-resolution.
The key component of FSRNet is the prior estimation network, which not only helps to improve the photometric recovery in terms of PSNR/SSIM, but also provides a solution for accurate geometry estimation directly from very LR images, as shown in the results of facial landmarks/parsing maps.
Extensive experimental results show that our FSRNet achieves superior performance than the state of the arts on unaligned face images, both quantitatively and qualitatively.
Following the main idea of this work, future research can be expanded in various aspects,
including designing a better prior estimation network, e.g., learning the fine SR network iteratively, and investigating other useful facial priors, e.g., texture.

\newpage
{\small
\bibliographystyle{ieee}
\bibliography{egbib}
}
\clearpage

\section{Appendix}

\maketitle
This supplementary material provides additional details of the following:

\noindent ($1$)  The network structure of the discriminator in FSRGAN.

\noindent ($2$) More qualitative comparisons with CBN~\cite{CBN_ECCV16} and several other state-of-the-art face SR methods~\cite{Jin_CVPR15,StructuredFH_CVPR13,NBF_ICCV15}.

\noindent ($3$) More visual examples to show robustness of our FSRNet and FSRGAN on different facial variations.

\subsection{Structure of Discriminator in FSRGAN}\label{section:1}
\begin{figure}[h!]
  \centering
  \includegraphics[trim={0 0 0 0mm},clip,width=0.95\linewidth]{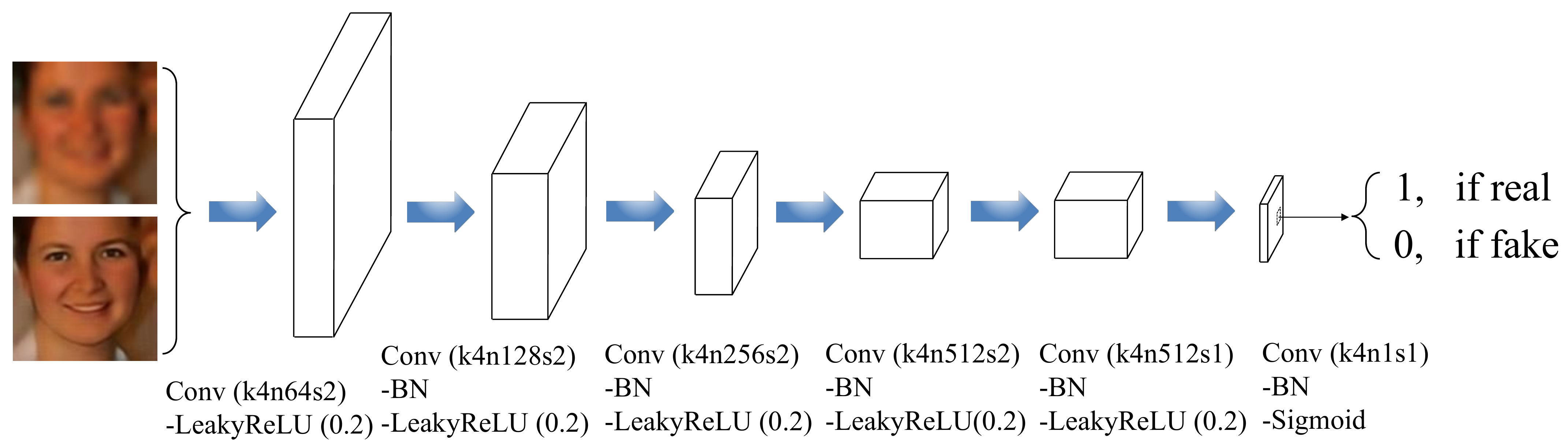}
  \caption{\small Structure of the discriminator network. The input is the concatenation of the low-resolution image with the recovered high-resolution image (fake) or the ground-truth one (real).} %
  \label{fig:structure of D}
\end{figure}

We follow the previous work~\cite{pix2pix_cvpr17} to use the ``PatchGAN'' structure in our discriminator, which down-samples the $128\times128$ input images to $8\times8$ feature maps, as shown in Fig.~\ref{fig:structure of D}. 
Each pixel in the feature map corresponds to a $16\times16$ patch in the original image, and the discriminator predicts the identity (fake or real) of each patch in the input.

\subsection{More Comparisons with State of the Arts}\label{section:2}
\begin{figure*}[h!]
  \centering
  \includegraphics[trim={0 0 0 0mm},clip,width=1.02\linewidth]{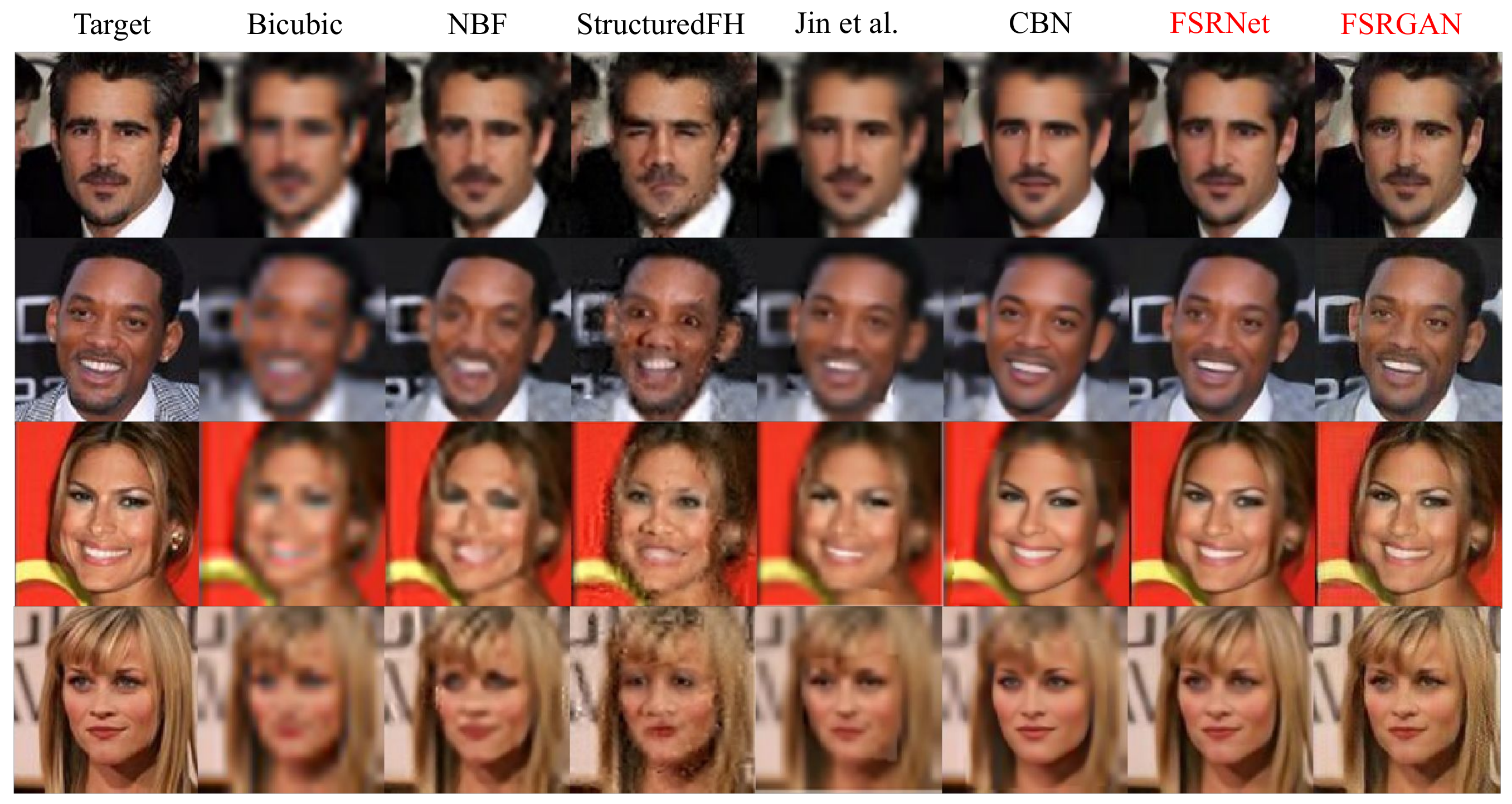}
 \vspace{-5mm}
  \caption{\small Qualitative results on dataset PubFig$83$.
  The test samples presented are imported directly from~\cite{CBN_ECCV16}.}
  \label{fig:more_comp_qualitative} %
\end{figure*}

Next, we present more qualitative comparisons with state-of-the-art face SR methods~\cite{CBN_ECCV16,Jin_CVPR15,StructuredFH_CVPR13,NBF_ICCV15}.
We follow the same experimental setting as CBN~\cite{CBN_ECCV16}, which uses the entire dataset celebA for training and test on dataset PubFig$83$~\cite{pubfig83}.
Here, we train our FSRNet/FSRGAN on the scale factor of $4\times$, with the first $201,599$ images or training and the last $1,000$ images for validation.
During testing, same as~\cite{CBN_ECCV16,Jin_CVPR15}, we blur HR faces with $\sigma=0.4$ to evaluate the robustness of our model on low-resolution and unknown gaussian blur simultaneously.
Results are shown in Fig.~\ref{fig:more_comp_qualitative}.
Compared with CBN, our models have $3$ advantages:
($1$) Our FSRNet looks more similar to the target image than CBN, and FSRGAN recovers competitive results to the target images.
($2$) There exists border effects in the recovered images of CBN, which is not a problem to our models.
($3$) CBN needs several steps and models to recover the HR image, which is slow and inconvenient. Our method only needs one forward process during inference, which is fast and convenient.

\begin{figure}[h!]
  \centering
  \includegraphics[trim={0 0 0 0mm},clip,width=1\linewidth]{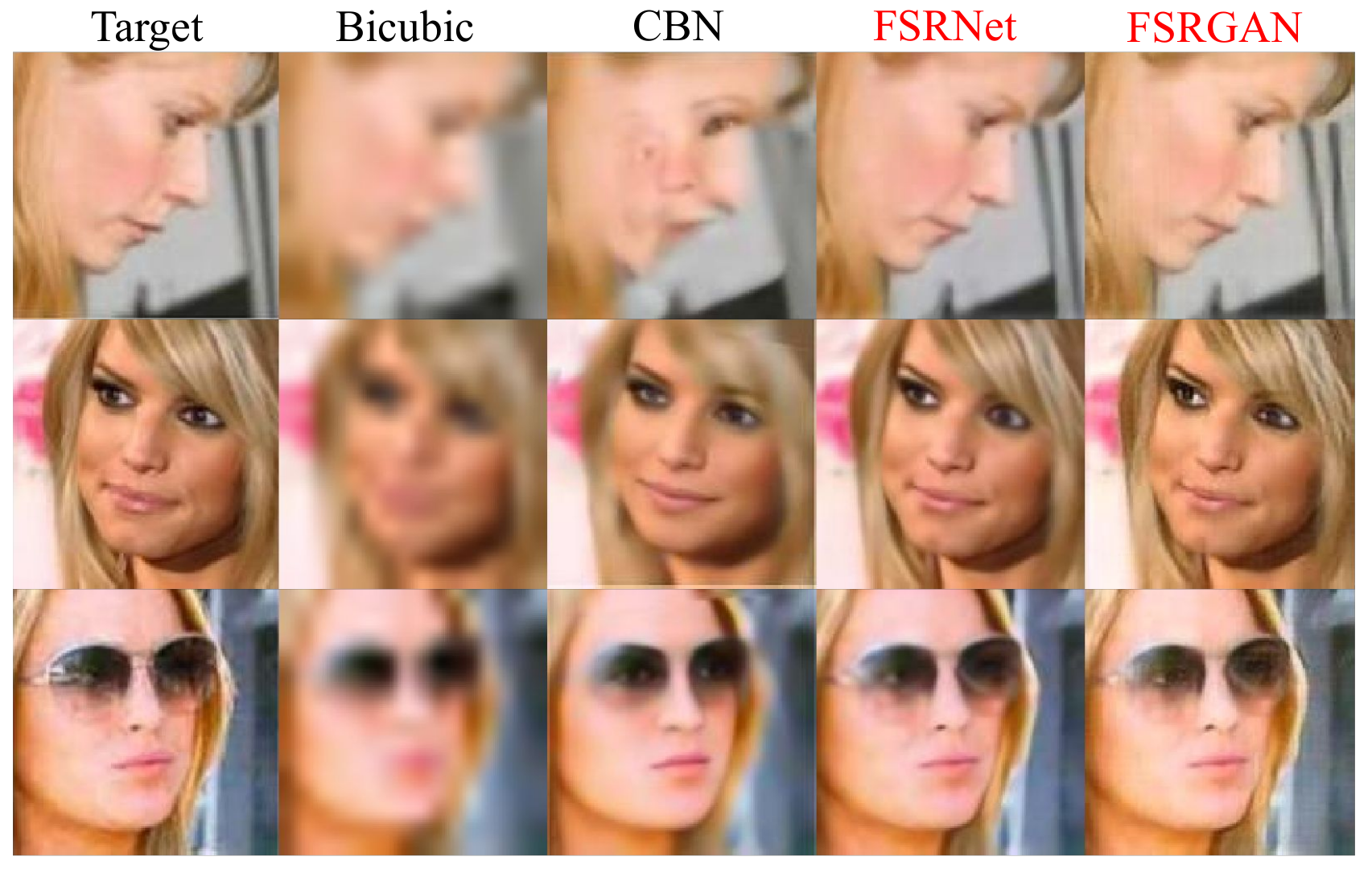}
 \vspace{-5mm}
  \caption{\small Qualitative results on the exact three representative failure cases of CBN~\cite{CBN_ECCV16}.
  Please zoom in to see the differences.}
  \label{fig:CBN_failure} \figvspace
\end{figure}

In Fig.~\ref{fig:CBN_failure}, we further present our results on recovering the exact three failure cases shown in CBN~\cite{CBN_ECCV16}.
Our models recover sharper and more accurate results than CBN in all three cases.
In the first example, CBN exists ghosting effect, while ours are more robust to facial misalignment and pose variations.
In the second example, CBN recovers incorrect gaze direction, while ours show the correct direction.
In the last example, over-synthesis of eyes is shown in CBN; ours also show this trend but are still better than CBN.

\begin{figure*}[tbp]
  \centering
  \includegraphics[trim={0 0 0 0mm},clip,width=1.02\linewidth]{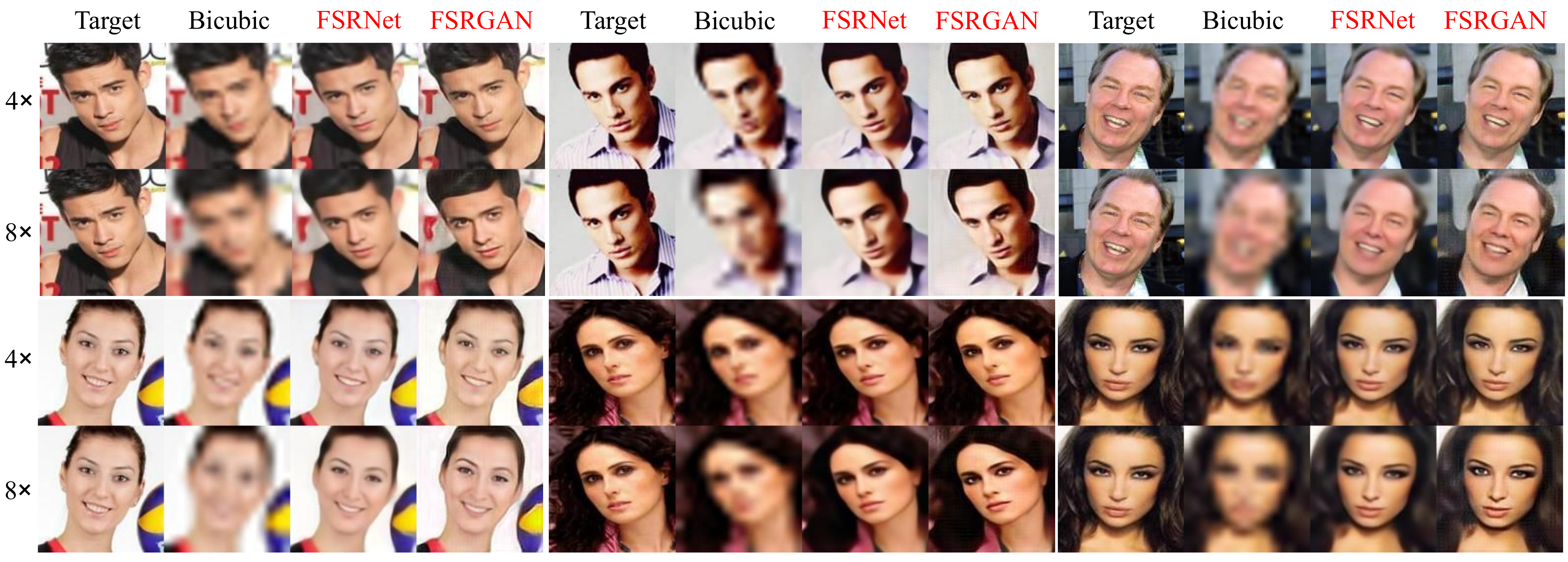}
 \vspace{-6mm}
  \caption{\small Robustness of FSRNet/FSRGAN to misalignment. Please zoom in for better view.}
  \label{fig:robustness_misalignment} \figvspace
\end{figure*}

\begin{figure*}[tbp]
  \centering
  \includegraphics[trim={0 0 0 0mm},clip,width=1.02\linewidth]{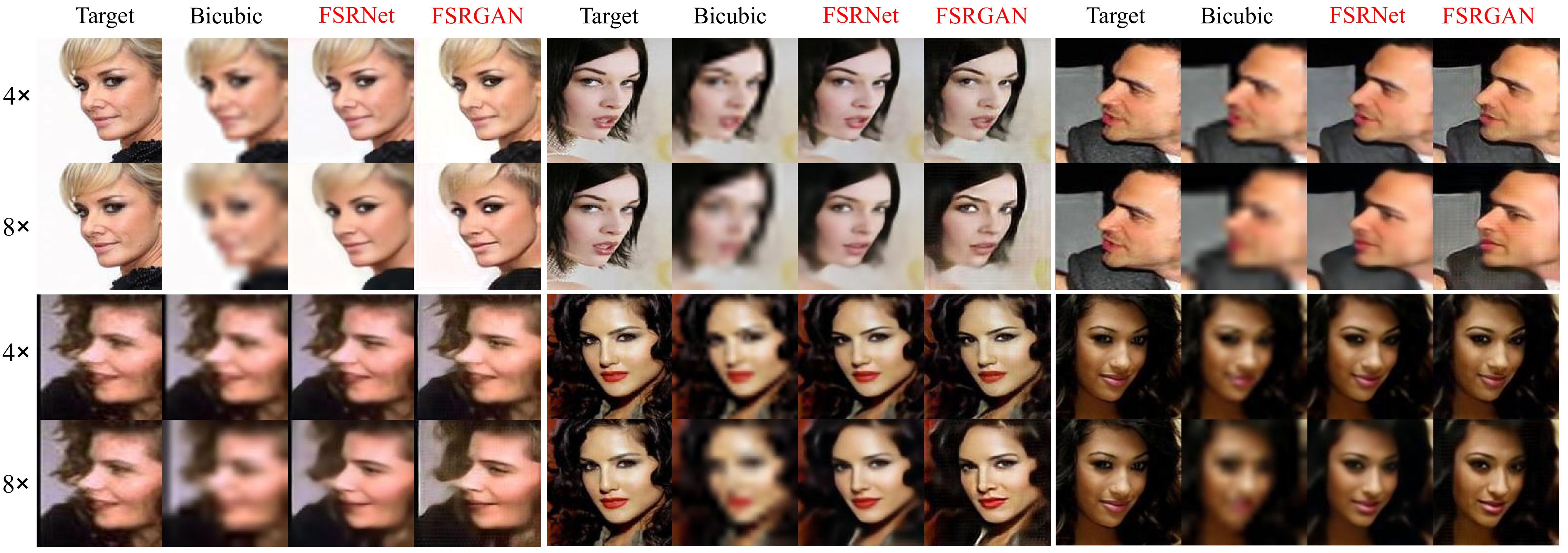}
 \vspace{-6mm}
  \caption{\small Robustness of FSRNet/FSRGAN to pose. Please zoom in for better view.}
  \label{fig:robustness_pose} \figvspace
\end{figure*}

\begin{figure*}[tbp]
  \centering
  \includegraphics[trim={0 0 0 0mm},clip,width=1.02\linewidth]{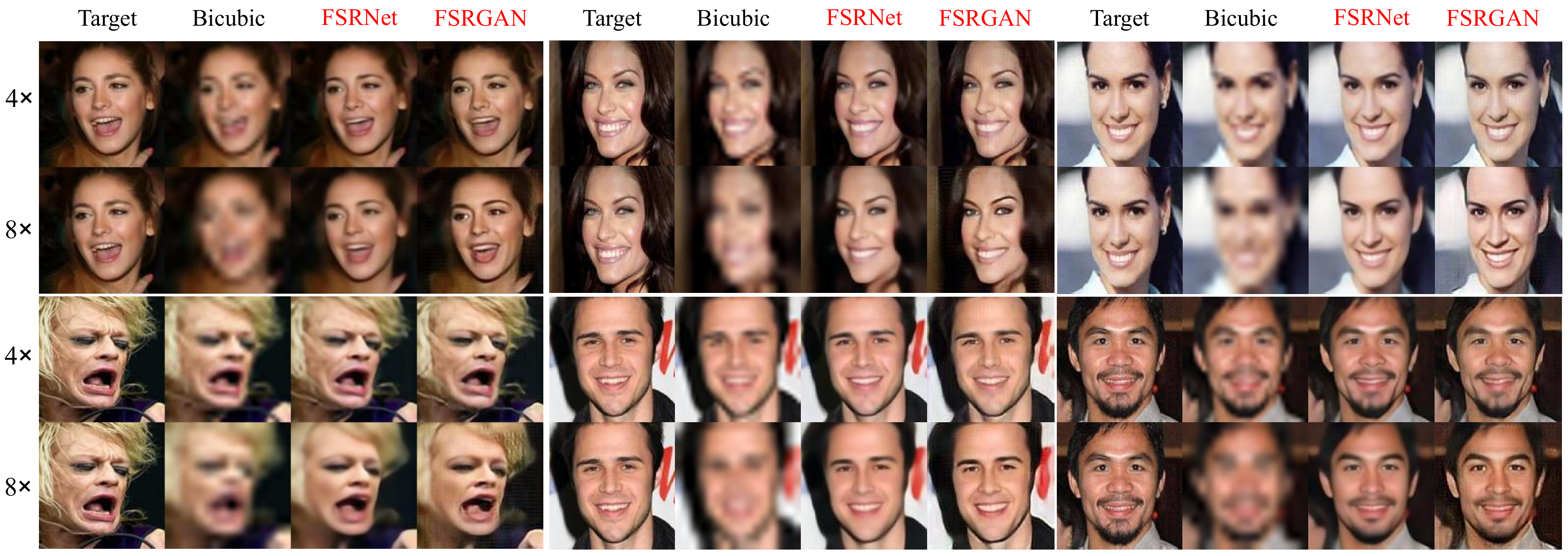}
 \vspace{-6mm}
  \caption{\small Robustness of FSRNet/FSRGAN to expression. Please zoom in for better view.}
  \label{fig:robustness_expression} \figvspace
\end{figure*}

\begin{figure*}[ht]
  \centering
  \includegraphics[trim={0 0 0 0mm},clip,width=1.02\linewidth]{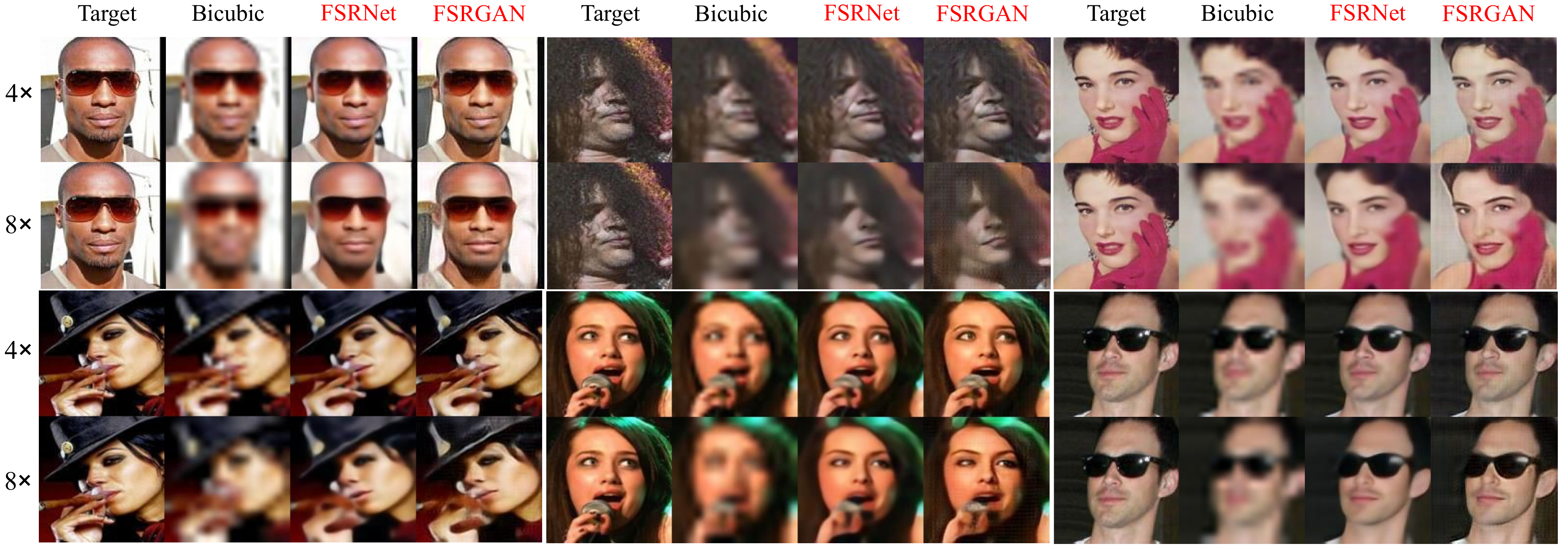}
 \vspace{-6mm}
  \caption{\small Robustness of FSRNet/FSRGAN to occlusion. Please zoom in for better view.}
  \label{fig:robustness_occlusion} \figvspace
\end{figure*}

\subsection{Robustness to Facial Variations}\label{section:3}
In our paper, we have shown the ability of our models on handling occluded faces.
Next, based on the model trained by $201,599$ images from celebA, we present more visual examples from the other $1,000$ validation images on two scale factors, $4\times$ and $8\times$, to show the robustness of our models on more facial variations.
Specifically, Fig.~\ref{fig:robustness_misalignment} shows the robustness of our models to misalignment;  Fig.~\ref{fig:robustness_pose} shows the robustness to pose; Fig.~\ref{fig:robustness_expression} shows the robustness to expression; and Fig.~\ref{fig:robustness_occlusion} shows the robustness to occlusions.
The extensive examples demonstrate the robustness of our models to different facial variations.
Last but not least, our models can recover extremely good results, which are indeed similar to the ground truth HR images, when handling scale factor to be $4\times$.

\end{document}